\documentclass[11pt,a4paper]{article}
\usepackage[hyperref]{acl2019}
\usepackage{times}
\usepackage{latexsym}

\usepackage{url}
\usepackage{hyperref}
\usepackage{url}

\usepackage[ruled,vlined]{algorithm2e}
\usepackage{algorithmic}
\usepackage{amsmath,amssymb,bbm,xcolor}
\usepackage{wrapfig,lipsum,booktabs}
\usepackage{natbib}
\usepackage{graphicx}
\usepackage{comment}
\usepackage{adjustbox}
\usepackage{graphicx}
\usepackage{amsmath,amssymb,enumerate,bbm,color,alltt}
\usepackage{hyperref}
\usepackage{url}
\usepackage{graphicx}
\usepackage{float}
\usepackage{stfloats}
\usepackage{subfig}
\usepackage{epstopdf}
\usepackage{longtable}
\usepackage{tabularx}
\usepackage{multirow}
\usepackage{booktabs}
\usepackage{subfig}
\usepackage{bigints}

\definecolor{myblue}{rgb}{0.158, 0.188, 0.978}
\definecolor{amber}{rgb}{1.0, 0.75, 0.0}
\definecolor{citrine}{rgb}{0.89, 0.82, 0.04}
\definecolor{carrotorange}{rgb}{0.93, 0.57, 0.13}
\definecolor{mypurple}{rgb}{0.558, 0.188, 0.678}
\definecolor{cadmiumgreen}{rgb}{0.0, 0.42, 0.24}
\definecolor{ao(english)}{rgb}{0.0, 0.5, 0.0}
\definecolor{cadmiumorange}{rgb}{0.93, 0.53, 0.18}
\definecolor{azure(colorwheel)}{rgb}{0.0, 0.5, 1.0}

\aclfinalcopy % Uncomment this line for the final submission
\title{Towards Generating Long and Coherent Text \\
    with Multi-Level Latent Variable Models} 
\author{Dinghan Shen \\
  Asli Celikyilmaz \\
  Yizhe Zhang \\
  Liqun Chen \\
  Xin Wang
  {\tt email@domain} \\\And
  Lawrence Carin \\
  Affiliation / Address line 1 \\
  Affiliation / Address line 2 \\
  Affiliation / Address line 3 \\
  {\tt email@domain} \\}

\author{Dinghan Shen$^{\mathbf{1}}$\thanks{~~This research was carried out during an internship at Microsoft Research.}, ~Asli Celikyilmaz$^{\mathbf{2}}$, ~Yizhe Zhang$^{\mathbf{2}}$, \smallskip 
	~  \\
\bf{~	Liqun Chen$^{\mathbf{1}}$, 	Xin Wang$^{\mathbf{3}}$, ~Jianfeng Gao$^{\mathbf{2}}$, ~Lawrence Carin$^{\mathbf{1}}$} \\
 \\
	$^{\mathbf{1}}$ Duke University~~~~
	$^{\mathbf{2}}$ Microsoft Research, Redmond~~~~ \\
	$^{\mathbf{3}}$ University of California, Santa Barbara \\
	{\tt dinghan.shen@duke.edu}
}

\date{}

\begin{document}
\maketitle
\begin{abstract}
Variational autoencoders (VAEs) have received much attention recently as an end-to-end architecture for text generation with latent variables. However, previous works typically focus on synthesizing relatively short sentences (up to $20$ words), and the posterior collapse issue has been widely identified in text-VAEs. In this paper, we propose to leverage several multi-level structures to learn a VAE model for generating \emph{long}, and \emph{coherent} text. In particular, a hierarchy of stochastic layers between the encoder and decoder networks is employed to abstract more informative and semantic-rich latent codes. Besides, we utilize a multi-level decoder structure to capture the coherent long-term structure inherent in long-form texts, by generating intermediate sentence representations as high-level \emph{plan vectors}. Extensive experimental results demonstrate that the proposed multi-level VAE model produces more coherent and less repetitive long text compared to baselines as well as can mitigate the
posterior-collapse issue. 

\end{abstract}

\section{Introduction}
% \vspace{-2mm}
The variational autoencoder (VAE) for text \citep{Bowman2016GeneratingSF} is a generative model in which a stochastic latent variable provides additional information to modulate the sequential text-generation process. 
VAEs have been used for various text processing tasks \citep{Semeniuta2017AHC, Zhao2017LearningDD, Kim2018SemiAmortizedVA, Du2018VariationalAD, hashimoto2018retrieve, Shen2018NASHTE, Xu2018SphericalLS,  Wang2019NAACL19}.
While most recent work has focused on generating relatively short sequences (\emph{e.g.,} a single sentence or multiple sentences up to around twenty words), generating long-form text (\emph{e.g.,} a single or multiple paragraphs) with deep latent-variable models has been less explored.

\begin{table}
	\begin{center}
		\vspace{2mm}
		\setlength{\tabcolsep}{4pt}
		\def\arraystretch{1.0}
		\begin{scriptsize}
			\begin{tabular} {p{1.4in}| p{1.4in}}
				\toprule[1.2pt]
				\textbf{\emph{flat}-VAE} (standard) & \textbf{\emph{multilevel}-VAE} (our model)\\
				\midrule[1.1pt]
				i went here for a grooming and a dog .
				it was very good .
				\textbf{\textcolor{azure(colorwheel)}{the owner is very nice and friendly} .
					\textcolor{azure(colorwheel)}{the owner is really nice and friendly}} .
				i don t know what they are doing . &
				i have been going to this nail salon for over a year now .
				the last time i went there .  
				the stylist was nice .  
				but the lady who did my nails . 
				she was very rude and did not have the best nail color i once had .
				\\
				\hline
				the staff is very friendly and helpful .
				\textbf{\textcolor{azure(colorwheel)}{the only reason i}} can t give them 5 stars .
				\textbf{\textcolor{azure(colorwheel)}{the only reason i}} am giving the ticket is because of the ticket . 
				\textbf{\textcolor{azure(colorwheel)}{can t help but}} the staff is so friendly and helpful .
				\textbf{\textcolor{azure(colorwheel)}{can t help but}} the parking lot is just the same .
				&
				i am a huge fan of this place . 
				my husband and i were looking for a place to get some good music . 
				this place was a little bit pricey .  
				but i was very happy with the service .
				the staff was friendly .
				\\
				\bottomrule[1.2pt]
			\end{tabular}
		\end{scriptsize}
		\vspace{-2mm}
		\caption{
		Comparison of samples generated from two generative models on the Yelp reviews dataset. The standard model struggles with repetitions of the same context or words (in \textcolor{azure(colorwheel)}{blue}), yielding non-coherent text. A hierarhical decoder with multi-layered latent variables eliminates redundancy and yields more coherent text planned around focused concepts.} 
% 		(See more examples in the supplementary material, Table~\ref{tab:hier_vae_c}).}
		\label{tab:hier_vae}
		\vspace{-3mm}
	\end{center}
\end{table}

Recurrent Neural Networks (RNNs) \citep{BahdanauICLR2015,ChopraNAACL2016} have mainly been used for most text VAE models \citep{Bowman2016GeneratingSF}. However, it may be difficult to scale RNNs for \emph{long-form} text generation, as they tend to generate text that is repetitive, ungrammatical, self-contradictory, overly generic and often lacking coherent long-term structure \citep{HoltzmanACL2018}. Two samples of text generated using standard VAE with an RNN decoder is shown in Table~\ref{tab:hier_vae}. 

In this work, we propose various multi-level network structures for the VAE model (\textit{ml}-VAE), to address coherency and repetitiveness challenges associated with long-form text generation. To generate globally-coherent long text sequences, it is desirable that both the \emph{higher-level} abstract features (\emph{e.g.}, topic, sentiment, etc.) and \emph{lower-level} fine-granularity details (\emph{e.g.}, specific word choices) of long text can be leveraged by the generative network. It's difficult for a standard RNN to capture such structure and learn to \emph{plan-ahead}. To improve the model's plan-ahead capability for capturing long-term dependency, following \citep{roberts2018hierarchical}, our first multi-level structure defines a hierarchical RNN decoder as the generative network that learns \emph{sentence-} and \emph{word-level} representations. Rather than using the latent code to initialize the RNN decoder directly, we found it more effective when first passing the latent code to a higher-level (sentence) RNN decoder, that outputs an embedding for the lower-level (word) RNN decoder that generates words. Since the low-level decoder network cannot fall back on autoregression, it gains a stronger reliance on the latent code to reconstruct the sequences.  
\vspace{-1mm}

Prior work has found that VAE training often suffers from \emph{“posterior collapse”}, in which the model ignores the latent code \citep{Bowman2016GeneratingSF}. This issue is related to the fact that the decoder network 
is usually parametrized with an autoregressive neural network, such as RNNs with teacher forcing scheme \citep{Bowman2016GeneratingSF, Yang2017ImprovedVA, Goyal2017ZForcingTS, Semeniuta2017AHC, Shen2018DeconvolutionalLM}.
Several strategies have been proposed (see optimization challenges in Section \ref{sec:opt}) to make the decoder less autoregressive, so
less \emph{contextual information} is utilized by the decoder network \citep{Yang2017ImprovedVA, Shen2018DeconvolutionalLM}. 
We argue that learning more informative latent codes can enhance the generative model without the need to lessen the contextual information. 
We propose leveraging a hierarchy of latent variables between the convolutional inference (encoder) networks and a multi-level recurrent generative network (decoder). With multiple stochastic layers, the prior of bottom-level latent variable is inferred from the data, rather than fixed as a standard Gaussian distribution as in typical VAEs \citep{kingma2013auto}. The induced \emph{latent code} distribution at the bottom level can be perceived as a Gaussian mixture, and thus is endowed with more flexibility to abstract meaningful features from the input text. 
While recent work has explored structures for more informative latent codes \cite{Kim2018SemiAmortizedVA, gu2018dialogwae}, \emph{ml}-VAE is
conceptually simple and easy to implement.
\vspace{-1mm}

We evaluate \emph{ml}-VAE on language modeling, unconditional and conditional text generation tasks. We show substantial improvements against several baseline methods in terms of
\emph{perplexity} on language modeling and quality of generated samples based on BLEU statistics and human evaluation.
% We show that our network can be generalized for conditional-generation scenarios.
% #####################################
\newcommand{\bs}[1]{\boldsymbol{#1}}
% \subsection{Variational Autoencoder (VAE)}
\vspace{-1mm}
\section{Variational Autoencoder (VAE)}
% #####################################
\vspace{-2mm}
Let $\bs{x}$ denote a text sequence, which consists of $L$ tokens, \emph{i.e.}, $x_1, x_2, ... , x_L$. 
A VAE encodes the text $\bs{x}$ using a recognition (encoder) model, $q_\phi(\bs{z}|\bs{x})$, parameterizing an approximate posterior distribution over a continuous latent variable $\bs{z}$ (whose prior is typically chosen as standard diagonal-covariance Gaussian). $\bs{z}$ is sampled stochastically from the posterior distribution, and text sequences $\bs{x}$ are generated conditioned on $\bs{z}$, via a generative (decoder) network, denoted as $p_{\bs{\theta}}(\bs{x}|\bs{z})$. 
A variational lower bound is typically used to estimate the parameters \citep{kingma2013auto}:
% \vspace{-2mm}
 \begin{align}
& \mathcal{L}_{\rm vae} = \mathbb{E}_{q_\phi(\bs{z}|\bs{x})} \left[\log \frac{p_\theta(\bs{x}|\bs{z})p(\bs{z})}{q_\phi(\bs{z}|\bs{x})}\right], \label{eq:vae} \\ %\notag \\
 & = \mathbb{E}_{q_\phi(\bs{z}|\bs{x})}[\log p_\theta(\bs{x}|\bs{z})] - D_{KL}(q_\phi(\bs{z}|\bs{x})||p(\bs{z})) , 
\vspace{-3mm} \notag
\end{align}
This lower bound is composed of a reconstruction loss (first term) that encourages the inference network to encode information necessary to generate the data and a KL regularizer (second term) to push $q_\phi(\bs{z}|\bs{x})$ towards the prior $p(\bs{z})$.

Although VAEs have been shown to be effective in a wide variety of text processing tasks (see related work), there are two challenges associated with generating longer sequences with VAEs: ($i$) they lack a long-term planning mechanism, which is critical for generating \emph{semantically-coherent} long texts \citep{Serdyuk2017TwinNU}; and 
($ii$) \emph{posterior collapse} issue. 
Concerning ($ii$), it was demonstrated in \citep{Bowman2016GeneratingSF} that due to the autoregressive nature of the RNN, the decoder tends to ignore the information from $\bs{z}$ entirely, resulting in an extremely small KL term (see Section~\ref{sec:opt}).

\begin{figure*}
	\centering
	\vspace{-1mm}
	\adjustbox{trim={0.0\width} {0.71\height} {0.\width} {0.01\height},clip}
    {
	\includegraphics[width=1.\textwidth]{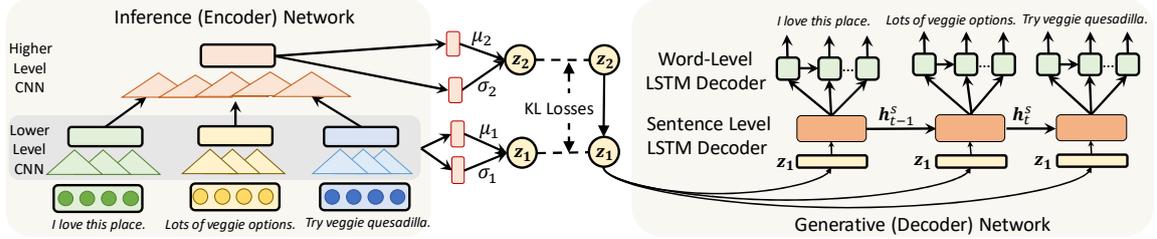}
    }
	\vspace{-2mm}
	\caption{The proposed \emph{multi-level} VAE with \emph{double} latent variables (\emph{ml}-VAE-D).}
	%Each point indicates one paper, and the color of each point indicates the 、\emph{category} it belongs to. }
	\vspace{-1mm}
	\label{fig:model}
\end{figure*}

% #####################################
\section{Multi-Level Generative Networks}
\vspace{-1mm}
% #####################################
\subsection{Single Latent Variable (\emph{ml}-VAE-S:)}
Our first multi-level model improves upon standard VAE models by introducing a
\emph{plan-ahead} ability to sequence generation. 
Instead of directly making word-level predictions only conditioned on the semantic information from $\bs{z}$, a series of \emph{plan vectors} are first generated based upon $\bs{z}$ with a \emph{sentence-level} LSTM decoder \cite{li2015hierarchical}.
Our hypothesis is that an explicit design of (inherently hierarchical) paragraph structure can capture sentence-level coherence and potentially mitigate repetitiveness.
Intuitively, when predicting each token, the decoder can use information from both the words generated previously and from sentence-level representations. 

An input paragraph consist of $M$ sentences, and each sentence $t$ has $N_t$ words, $t$=$1$,$\dots,M$.
To generate the plan vectors, the model first samples a latent code $\bs{z}$ through a one-layer multi-layered perceptron (MLP), with ReLU activation functions, to obtain the starting state of the sentence-level LSTM decoder.
Subsequent sentence representations, namely the \emph{plan vectors}, are generated with the sentence-level LSTM in a sequential manner:
 \begin{align}
\vspace{-3mm}
\bs{h}_t^s = \text{LSTM}^{\text{sent}}(\bs{h}_{t-1}^s, \bs{z}), %\,\text{for}\,\,t = 1, 2, 3, ..., M 
\vspace{-3mm} %\notag 
\end{align}
The latent code $\bs{z}$ can be considered as a paragraph-level abstraction, relating to information about the semantics of each generated subsequence.
Therefore we input $\bs{z}$ at each time step of the sentence-level LSTM, to predict the sentence representation. Our single-latent-variable model sketched in Figure~\ref{fig:model_s} of supplementary material.

The generated sentence-level plan vectors are then passed onto the word-level LSTM decoder to generate the words for each sentence. To generate each word of a sentence $t$, the corresponding \emph{plan vector}, $\bs{h}_t^s$, is concatenated with the word embedding of the previous word and fed to LSTM$^{word}$ at every time step \footnote{We use teacher-forcing during training and greedy decoding at test time.}. Let $w_{t, i}$ denote the $i$-th token of the $t$-th sentence  This process can be expressed as (for $t = 1, 2, ..., M$ and $i = 1, 2, 3, ..., N_t$):
\begin{align}
\bs{h}_{t, i}^w = \text{LSTM}^{\text{word}}(&\bs{h}_{t, i-1}^w, \bs{h}_{t}^s, \bs{W_e}[w_{t, i-1}]), \, \,\\
p(w_{t, i}|{w}_{t, <i}, \bs{h}_{t}^s) &= \text{softmax}(\bs{V} \bs{h}_{t, i}^w),
\end{align}
The initial state $\bs{h}_{t, 0}^w$ of LSTM$^{word}$ is inferred from the corresponding plan vector via an MLP layer. $\bs{V}$ represents the weight matrix for computing distribution over words, and $\bs{W_e}$ are word embeddings to be learned. For each sentence, once the special \texttt{\_END} token is generated, the word-level LSTM stops decoding \footnote{Each sentence is padded with an \texttt{\_END} token.}. LSTM$^{word}$ decoder parameters are shared for each generated sentence. 

% #####################################
\vspace{-1mm}
\subsection{Double Latent Variables (\emph{ml}-VAE-D):}
% #####################################
Similar architectures of our single latent variable \emph{ml}-VAE-S model have been applied recently for multi-turn dialog response generation \cite{Serban2017AHL, Park2018AHL}, mainly focusing on short (one-sentence) response generation. 
Different from these works, our goal is to generate \emph{long text} which introduces additional challenges to the hierarchical generative network. 
We hypothesize that with the two-level LSTM decoder embedded into the VAE framework, the load of capturing global and local semantics are handled differently than the \emph{flat}-VAEs \citep{chen2016variational}. While the multi-level LSTM decoder can capture relatively detailed information (\emph{e.g.}, word-level (local) coherence) via the word- and sentence-level LSTM networks, the latent codes of the VAE are encouraged to abstract more global and high-level semantic features of multiple sentences of long text.

Our double latent variable extension, \emph{ml}-VAE-D, is shown in Figure~\ref{fig:model}.
The inference network encodes upward through each latent variable to infer their posterior distributions, while the generative network samples downward to obtain the distributions over the latent variables. 
The distribution of the latent variable at the bottom is inferred from the top-layer latent codes, rather than fixed (as in a standard VAE model). This also introduces flexibility to the model to abstract useful high-level features \citep{gulrajani2016pixelvae}, which can then be leveraged by the multi-level LSTM network.
Without loss of generality, here we choose to employ a two-layer hierarchy of latent variables, where the bottom and top layers are denoted as $\bs{z_1}$ and $\bs{z_2}$, respectively, which can be easily extended to multiple latent-variable layers. 

Another important advantage of multi-layer latent variables in the VAE framework is related to the \emph{posterior collapse} issue. Even though the single latent variable model (\emph{ml}-VAE-S) defines a multi-level LSTM decoder, the posterior collapse can still exist since the LSTM decoder can still ignore the latent codes due to its autoregressive property. With the hierarchical latent variables, we propose a novel strategy to mitigate this problem, by making less restrictive assumptions regarding the prior distribution of the latent variable. Our model yields a larger KL loss term relative to \emph{flat}-VAEs, indicating more informative latent codes.

The posterior distributions over the latent variables are assumed to be conditionally independent given the input $x$. 
We can represent the joint posterior distribution of the two latent variables as \footnote{We assume $\bs{z_1}$ and $\bs{z_2}$ to be independent on the encoder side, since this specification will yield a closed-form expression for the KL loss between $p_{\bs {\theta}}(\bs{z_1}, \bs{z_2})$ and $q_{\bs {\phi}}(\bs{z_1}, \bs{z_2}|\bs{x})$.}: 
\vspace{-2mm}
\begin{align} \label{eq:prior}
q_{\bs {\phi}}(\bs{z_1}, \bs{z_2}|\bs{x}) = q_{\bs {\phi}}(\bs{z_2}|\bs{x})q_{\bs {\phi}}(\bs{z_1}|\bs{x})
\end{align}
Concerning the generative network, the latent variable at the bottom is sampled conditioned on the one at the top. Thus, we have:
\begin{align} \label{eq:posterior}
p_{\bs {\theta}}(\bs{z_1}, \bs{z_2}) = p_{\bs {\theta}}(\bs{z_2})p_{\bs {\theta}}(\bs{z_1}|\bs{z_2})
\end{align}
%
% To optimize the parameters of the inference and generative networks, 
$D_{KL}(q_\phi(\bs{z}|\bs{x})||p(\bs{z}))$, the second term of the VAE objective, then becomes the KL divergence between joint posterior and prior distributions of the two latent variables. Under the assumptions of (\ref{eq:prior}) and (\ref{eq:posterior}), the variational lower bound yields: 
\begin{align}
&\mathcal{L}_{\rm vae}    = \mathbb{E}_{q(\bs{z_1}|\bs{x})} [\log p(\bs{x}|\bs{z_1})] \notag \\
& \quad \quad - D_{\text{KL}}(q(\bs{z_1}, \bs{z_2}|\bs{x})||p(\bs{z_1},\bs{z_2}))    \label{eq:elbo_1} 
\end{align}
Abbreviarting $p_{\bs {\theta}}$ and $q_{\bs {\phi}}$ with $p$ and $q$, we get:
% \yz{two =}
\begin{align}
& D_{\text{KL}}(q(\bs{z_1}, \bs{z_2}|\bs{x})||p(\bs{z_1},\bs{z_2})) \notag \\
% & D_{\text{KL}}
&=\int q(\bs{z_2}|\bs{x})q(\bs{z_1}|\bs{x})\log\frac{q(\bs{z_2}|\bs{x})q(\bs{z_1}|\bs{x}) }{p(\bs{z_2})p(\bs{z_1}|\bs{z_2})} d \bs{z_1}d \bs{z_2} \notag \\
& =  \int_{\bs{z_1}, \bs{z_2}} [q_{\bs {\phi}}(\bs{z_2}|\bs{x})q_{\bs {\phi}}(\bs{z_1}|\bs{x}) \log \frac{q_{\bs {\phi}}(\bs{z_1}|\bs{x}) }{p_{\bs {\theta}}(\bs{z_1}|\bs{z_2})} \notag \\ 
&\quad \quad +  q(\bs{z_2}|\bs{x})q(\bs{z_1}|\bs{x}) \log\frac{q(\bs{z_2}|\bs{x})}{p(\bs{z_2})}] d \bs{z_1}d \bs{z_2} \notag \\
& =\mathbb{E}_{q(\bs{z_2}|\bs{x})}[D_{KL}(q(\bs{z_1}|\bs{x})||p(\bs{z_1}|\bs{z_2}))] \notag \\
& \quad \quad + D_{KL}(q(\bs{z_2}|\bs{x})||p(\bs{z_2}))  \label{eq:elbo_2} 
\vspace{-3mm}
\end{align}
The left-hand side of (\ref{eq:elbo_2}) is the abbreviation of $D_{\text{KL}}(q_\phi(\bs{z_1}, \bs{z_2}|\bs{x})||p(\bs{z_1}, \bs{z_2}))$. Given the Gaussian assumption for both the prior and posterior distributions, both KL divergence terms can be written in closed-form.

% #####################################
% \subsection{Model Specifications}
% #####################################
To abstract meaningful representations from the input paragraphs, we choose a hierarchical CNN architecture for the inference/encoder networks for both single and double latent variable models.
We use sentence-level CNNs to encode each sentence into a fixed-length vector, which are then aggregated and send to paragraph-level CNN encoder. The inference networks parameterizing $q(\bs{z_1}|\bs{x})$ and $q(\bs{z_2}|\bs{x})$ share the same parameters as the lower-level CNN. 

The single-variable \emph{ml}-VAE-S model feeds the paragraph feature vector into the linear layers to infer the mean and variance of the latent variable $z$.
In the double-variable model {\emph{ml}-VAE-D}, the feature vector is transformed with two MLP layers, and then is used to compute the mean and variance of the top-level latent variable. 
\section{Related Work}
%\yz{add some reference for text generation with VAE or GAN?
\paragraph{\textbf{VAE for text generation.}}  VAEs trained under the neural variational inference (NVI) framework, has been widely used for generating text sequences: \citep{Bowman2016GeneratingSF, Yang2017ImprovedVA, Semeniuta2017AHC,miao2016neural, Serban2017AHL, Miao2017DiscoveringDL,  Zhao2017LearningDD, Shen2017ACV, 
Guu2018GeneratingSB, Kim2018SemiAmortizedVA, Yin2018StructVAETL, Kaiser2018FastDI,  Bahuleyan2018VariationalAF, Chen2018VariationalSL, Deng2018LatentAA, shah2018generative}.

By encouraging the latent feature space to match a prior distribution within an encoder-decoder architecture, the learned latent variable could potentially encode high-level semantic features and serve as a global representation during the decoding process \citep{Bowman2016GeneratingSF}.
The generated results are also endowed with better diversity due to the sampling procedure of the latent codes \citep{Zhao2017LearningDD}. 
Generative Adversarial Networks (GANs) \citep{Yu2017SeqGANSG, Hu2017TowardCG, Zhang2017AdversarialFM, fedus2018maskgan, chen2018adversarial}, is another type of generative models that are commonly used for text generation. 
However, existing works have mostly focused on generating one sentence (or multiple sentences with at most twenty words in total).
The task of generating relatively longer units of text has been less explored.

% \vspace{-2mm}
\paragraph{\textbf{Optimization Challenges.}} 
\label{sec:opt}
% \vspace{-2mm}
The ``posterior collapse'' issue associated with training text-VAEs was first outlined by \citep{Bowman2016GeneratingSF}. They used two strategies, \textit{KL divergence annealing} and \textit{word dropout}, however, none of them help to improve the perplexity compared to a plain neural language model.
\cite{Yang2017ImprovedVA} argue that the small KL term relates to the strong autoregressive nature of an LSTM generative network, and they proposed to utilize a dilated CNN as a decoder to improve the informativeness of the latent variable. 
\citep{zhao2018unsupervised} proposed to augment the VAE training objective with an additional mutual information term.  
This yields an intractable integral in the case where the latent variables are continuous. Recent work \cite{laggingVAE2019,cyclical2019} has shown that advanced scheduling can mitigate the posterior collapse issue. We instead introduce more flexible priors and hierarchical encoder and decoder structures to deal with posterior collapse.

\paragraph{Hierarchical Structures.} 
\vspace{-1.5mm}
Natural language is inherently hierarchical (characters form a word, words form a sentence, sentences form a paragraph, paragraphs from a document, \emph{etc}.). Previous work used multi-level LSTM encoders \cite{yang2016hierarchical} or hierarchical autoencoders \cite{Li2015AHN} to learn hierarchical representations for long text or defined a stochastic latent variable for each sentence at decoding time \citep{Serban2017AHL}. In contrast, our model encodes the entire paragraph into one \emph{single} latent variable. The latent variable learned in our model relates more to the global semantic information of a paragraph, whereas those in \citep{Serban2017AHL} mainly contain the local information of a specific sentence.
\citeauthor{Park2018AHL}\cite{Park2018AHL} introduced a variational hierarchical conversational model (VHCR) with global and local latent variables. They generate local/utterance variables condintioned on the global latent variable, assuming standard dialog-covariance Gaussian for both latent variables. In contrast, both our latent variables in \emph{ml}-VAE-D are designed to contain global information. \emph{ml}-VAE learns the prior of the bottom-level latent variable from the data, yielding more flexible prior relative to a fixed prior and promising results in mitigating the issue of ``posterior collapse" in the experiments.
The responses in VHCR are generated conditionally on the latent variables and context, while our \emph{ml}-VAE-D model captures the underlying data distribution of the entire paragraph in the bottom latent variable ($\bs{z_1}$), so the global latent variable contains more information. 

%###################################
% \vspace{-2mm}
\section{Experiments}
% \vspace{-1mm}
%###################################
\subsection{Experimental Setup}
% \vspace{-2mm}
\paragraph{Datasets} We conducted experiments on both generic (unconditional) long-form text
generation and conditional paragraph generation (with additional text input as auxiliary information). For the former, we use two datasets: Yelp Reviews \citep{zhang2015character}
and arXiv Abstracts. For the conditional-generation experiments, we consider the task of synthesizing a paper abstract conditioned on the paper title (with the arXiv Abstracts dataset)\footnote{Our goal is to analyze if the proposed architecture can discover different concepts with the hierarchical decoding and latent code structures, thus we use the arxiv dataset with indicated domains for demonstration purposes. We leave the common summarization datasets for future research.}. 
Details on dataset statistics and model architectures are provided in the supplementary material.
% \vspace{-1mm}
\paragraph{Baselines} We implement the following langauge modeling baselines: language model with a flat LSTM decoder (\emph{flat}-LM), VAE with a flat LSTM decoder (\emph{flat}-VAE),
%(similar to the model proposed in \citep{Bowman2016GeneratingSF}), 
and language model with a multi-level LSTM decoder (\emph{ml}-LM)\footnote{We only experimented with state of the art models with similar architectures to our models, since our goal is to investigate the impact of hiararhical VAE structure on the text generation. More efficient new encoder and decoder architectures such as non-autoregressive models is a direction for extending this work.}.

For generic text generation, we build models using two recently proposed generative models as baselines: Adversarial Autoencoders (AAE) \citep{makhzani2015aae} and Adversarially-Regularized Autoencoders (ARAE) \citep{Zhao2018AdversariallyRA}. Instead of penalizing the KL divergence term, AAE introduces a discriminator network to match the prior and posterior distributions of the latent variable. AARE model extends AAE by introducing Wassertein GAN loss \cite{WassersteinGAN2017} and a stronger generator network. We build two variants of our multi-level VAE models: single latent variable \emph{ml}-VAE-S and double latent variable \emph{ml}-VAE-D. 
%Our code can be downloaded here\footnote{http://github.com/???}.
Our code will be released to encourage future research.
\vspace{-2mm}
\subsection{Language Modeling Results}
\vspace{-1.5mm}
We report negative log likelihood (NLL) and perplexity (PPL) results on Yelp and arXiv datasets.
Following \citep{Bowman2016GeneratingSF,Yang2017ImprovedVA,Kim2018SemiAmortizedVA}, we use the KL loss term to measure the extent of \textit{``posterior collapse''}.

As shown in Table~\ref{tab:nll}, the standard \emph{flat}-VAE on Yelp dataset yields a KL divergence term very close to zero, indicating that the generative model makes negligible use of the information from latent variable $\bs{z}$. The \emph{flat}-VAE model obtains slightly worse NNL and PPL relative to a flat LSTM-based language model. With multi-level LSTM decoder, our \emph{ml}-VAE-S yields increased KL divergence, demonstrating that the VAE model tends to leverage more information from the latent variable in the decoding stage. The PPL of \emph{ml}-VAE-S is also decreased from 47.9 to 46.6 (compared to \emph{ml}-LM), indicating that the sampled latent codes improve word-level predictions.
\begin{table}
	\centering
	%\vskip 0.0in
	\def\arraystretch{1.2}
	\small
	\setlength{\tabcolsep}{2.65pt}
	\vspace{2mm}
	\begin{tabular} {ccccccc}
		%\hline
		\toprule[1.2pt]
		\multirow{2}{*}{\textbf{Model}} & \multicolumn{3}{c}{\textbf{Yelp}} & \multicolumn{3}{c}{\textbf{arXiv}} \\
		\cmidrule(l){2-4} \cmidrule(l){5-7}
		& \textbf{NLL} & \textbf{KL} & \textbf{PPL}  &  	\textbf{NLL} & \textbf{KL} & \textbf{PPL} \\
		%\textbf{Model} &  	\textbf{NLL} & \textbf{KL} & \textbf{PPL}  &  	\textbf{NLL} & \textbf{KL} & \textbf{PPL}  \\
		\midrule[1.1pt]
		\emph{flat}-LM       & 162.6 & - & 48.0  & 218.7  & - & 57.6  \\
		\emph{flat}-VAE     & $\leq$ 163.1 & 0.01 & $\leq$ 49.2  & $\leq$ 219.5  & 0.01 & $\leq$ 58.4  \\
		\hline 
		\emph{ml}-LM       & 162.4 & - & 47.9  & 219.3 & - & 58.1  \\
		\emph{ml}-VAE-S      & $\leq$ 160.8  & 3.6 & $\leq$ 46.6  & $\leq$ 216.8 & 5.3 & $\leq$ 55.6   \\
		\emph{ml}-VAE-D     & $\leq$ \textbf{160.2} & 6.8 & $\leq$ \textbf{45.8} & $\leq$ \textbf{215.6} & 12.7 & $\leq$ \textbf{54.3} \\
		\bottomrule[1.2pt]
		%\hline
	\end{tabular}
	\vspace{-2mm}
	\caption{Language modeling results on Yelp and arXiv data. Upper block are baselines, and lower are our models.}
	\label{tab:nll}
	%\end{scriptsize}
	\vspace{-4mm}
\end{table}

Our double latent variable model, \emph{ml}-VAE-D, exhibits an even larger KL divergence cost term (increased from $3.6$ to $6.8$) than single latent variable model, indicating that more information from the latent variable has been utilized by the generative network. This may be due to the fact that the latent variable priors of the \emph{ml}-VAE-D model are inferred from the data, rather than a fixed standard Gaussian distribution. As a result, the model is endowed with more flexibility to encode informative semantic features in the latent variables, yet matching their posterior distributions to the corresponding priors. \emph{ml}-VAE-D achieves the best PPL results on both
datasets (on the arXiv dataset, our hierarchical decoder outperforms the \emph{ml}-LM by
reducing the PPL from $58.1$ down to $54.3$).
\subsection{Unconditional Text Generation}
% \vspace{-2mm}
We evaluate the quality of generated paragraphs as follows.
We randomly sample $1000$ latent codes and send them to all trained generative models to generate text. 
We use corpus-level BLEU score \citep{papineni2002bleu} to quantitatively evaluate the generated paragraphs. 
Following strategy in \citep{Yu2017SeqGANSG, Zhang2017AdversarialFM} we use the entire test set as the reference for each generated text, and get the average BLEU scores\footnote{Being interested in longer text generation, we evaluate our models on the n-gram reconscturion ability (where n$>$1).} over $1000$ generated sentences for each model.

The results are in Table~\ref{tab:bleu}. VAE tends to be a stronger baseline for paragraph generation, exhibiting higher corpus-level BLEU scores than both AAE and ARAE. This observation is consistent with the results in \citep{cifka2018eval} in Table~\ref{tab:bleu}.
The VAE with multi-level decoder demonstrates better BLEU scores than the one with a flat decoder, indicating that the plan-ahead mechanism associated with the hierarchical decoding process indeed benefits the sampling quality. \emph{ml}-VAE-D exhibits slightly better results than \emph{ml}-VAE-S. We attribute this to the more flexible prior distribution of \emph{ml}-VAE-D, which improves the ability of inference networks to extract semantic features from a paragraph, yielding more informative latent codes. 

\begin{figure} %{R}{1.66in}
	\centering
	\vspace{0mm}
	\includegraphics[width=0.43\textwidth]{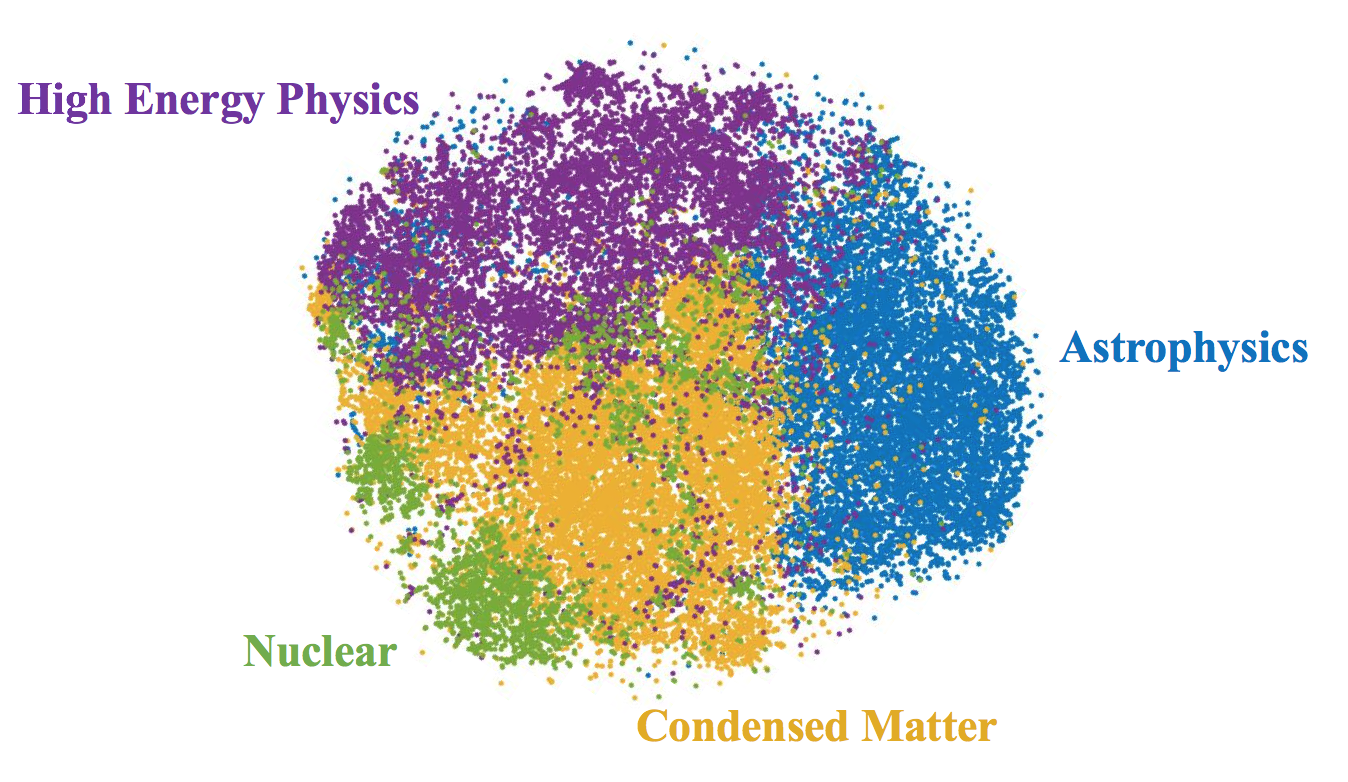}
	\vspace{-4mm}
	\caption{\emph{t}-SNE visualization of the learned latent codes. }
	%Each point indicates one paper, and the color of each point indicates the 、\emph{category} it belongs to. }
	\vspace{-2mm}
	\label{fig:tsne}
	%\end{table}
\end{figure}

We visualize the learnt latent variables to analyze if our models can extract global features. Using the arXiv dataset, we select the most frequent four article topics and re-train our \emph{ml}-VAE-D model on the corresponding abstracts in an unsupervised way (no topic information is used).
We sample bottom-level latent codes from the learned model and plot them with \emph{t}-SNE in Figure~\ref{fig:tsne}. Each point indicates one paper abstract and the color of each point indicates the topic it belongs to. The embeddings of the same label are very close in the 2-D plot, while those with different labels are relatively farther away from each other. The embeddings of the \emph{High Energy Physics} and \emph{Nuclear} topic abstracts are meshed, which is expected since these two topics are semantically highly related. The inference network can extract meaningful global patterns from the input paragraph.

In Table~\ref{tab:hier_vae} two samples of generations from \emph{flat}-VAE and \emph{ml}-VAE-D are shown. Compared to our hierarchical model, a flat decoder with a flat VAE exibits repetitions as well as suffers from uninformative sentences. The hierarchical model generates reviews that contain more information with less repetitions (word or semantic semantic repetitions), and tend to be semantically-coherent.

\begin{table}[t]
	\centering
	%\vskip 0.0in
	\def\arraystretch{1.2}
	\setlength{\tabcolsep}{4pt}
	%\footnotesize 
	\small 
	\vspace{2mm}
	\begin{tabular}{cc@{\hskip5pt}c@{\hskip5pt}c|@{\hskip5pt}c@{\hskip5pt}c@{\hskip5pt}cc}
		%\hline
		\toprule[1pt]
		\multirow{2}{*}{\textbf{Model}} & \multicolumn{3}{c}{\textbf{Yelp}} & \multicolumn{3}{c}{\textbf{arXiv}}
		\\
		\cmidrule(l){2-4} \cmidrule(l){5-7}
		& \textbf{B-2} & \textbf{B-3} & \textbf{B-4} & \textbf{B-2} & \textbf{B-3} & \textbf{B-4}  \\
		\toprule[1pt]
		%\textbf{Model} &  	\textbf{BLEU-2} & \textbf{BLEU-3} & \textbf{BLEU-4} & \textbf{BLEU-5} \\ 
		\emph ARAE      & 0.684 & 0.524 & 0.350  & 0.624 & 0.475 & 0.305    \\
		\emph AAE      & 0.735 & 0.623 & 0.383  & 0.729 & 0.564 & 0.342     \\
		\emph{flat}-VAE     & 0.855 & 0.705 & 0.515  & 0.784 & 0.625 & 0.421    \\
		\hline
		\emph{ml}-VAE-S     & 0.901 & 0.744 & 0.531 & 0.821 & \textbf{0.663} & 0.447    \\
		\emph{ml}-VAE-D   & \textbf{0.912} & \textbf{0.755} & \textbf{0.549} & \textbf{0.825} & 0.657 & \textbf{0.460}    \\
		\bottomrule[1.2pt]
		%\hline
	\end{tabular}
% 	\usepackage{caption}
% \vspace{-2mm}
% \captionsetup{width=.90\textwidth}
            \caption{Evaluation results for generated sequences by our models and baselines on corpus-level BLEU scores (\textbf{B-n} denotes the corpus-level \emph{BLEU-n} score.)}
	\label{tab:bleu}
% 	\vspace{-2mm}
\end{table}

\begin{comment}
\begin{table}
	\centering
	%\vskip 0.0in
	\def\arraystretch{1.0}
	\setlength{\tabcolsep}{3pt}
	%\footnotesize 
	\small 
	\vspace{-3mm}
	\begin{tabular}{cc|@{\hskip2pt}c|@{\hskip2pt}c|@{\hskip2pt}c|@{\hskip3pt}c|@{\hskip2pt}c|@{\hskip2pt}c|@{\hskip2pt}c}
		%\hline
		\toprule[1pt]
		\multirow{2}{*}{\textbf{Model}} & \multicolumn{4}{c}{\textbf{Yelp}} & \multicolumn{4}{c}{\textbf{arXiv}}
		\\
		\cmidrule(l){2-5} \cmidrule(l){6-9}
		& \textbf{B-2} & \textbf{B-3} & \textbf{B-4}  &  	\textbf{B-5} & \textbf{B-2} & \textbf{B-3} & \textbf{B-4}  &  	\textbf{B-5}  \\
		\toprule[1pt]
		%\textbf{Model} &  	\textbf{BLEU-2} & \textbf{BLEU-3} & \textbf{BLEU-4} & \textbf{BLEU-5} \\ 
		\emph ARAE      & 0.684 & 0.524 & 0.350 & 0.104  & 0.624 & 0.475 & 0.305 & 0.124   \\
		\emph AAE      & 0.735 & 0.623 & 0.383 & 0.167  & 0.729 & 0.564 & 0.342 & 0.153    \\
		\emph{flat}-VAE     & 0.855 & 0.705 & 0.515 & 0.330   & 0.784 & 0.625 & 0.421 & 0.247   \\
		\hline
		\emph{ml}-VAE-S     & 0.901 & 0.744 & 0.531 & 0.336 & 0.821 & \textbf{0.663} & 0.447 & 0.273   \\
		\emph{ml}-VAE-D   & \textbf{0.912} & \textbf{0.755} & \textbf{0.549} & \textbf{0.347} & \textbf{0.825} & 0.657 & \textbf{0.460} & \textbf{0.282}    \\
		\bottomrule[1.2pt]
		%\hline
	\end{tabular}
% 	\usepackage{caption}
    \vspace{-2mm}
% \captionsetup{width=.90\textwidth}
            \caption{\small Evaluation of generated sequences by our models and baselines on corpus-level BLEU scores (\textbf{B-n} denotes the corpus-level \emph{BLEU-n} score.).}
	\label{tab:bleu}
 	\vspace{-2mm}
\end{table}
\end{comment}

\vspace{-2mm}
\paragraph{Diversity of Generated Paragraphs} We evaluate the diversity of random samples from a trained model, since one model might generate realistic-looking sentences while suffering from severe mode collapse (\emph{i.e.}, low diversity). We use three metrics to measure the diversity of generated paragraphs: Self-BLEU scores \citep{zhu2018texygen}, unique $n$-grams \citep{fedus2018maskgan} and the entropy score \citep{zhang2018generating}. For a set of sampled sentences, the Self-BLEU metric is the BLEU score of each sample with respect to all other samples as the reference (the numbers over all samples are then averaged); the unique score computes the percentage of \emph{unique} $n$-grams within all the generated reviews; and the entropy score measures how evenly the empirical $n$-gram distribution is for a given sentence, which does not depend on the size of testing data, as opposed to unique scores. In all three metrics, lower is better.
We randomly sample $1000$ reviews from each model.

The results are shown in Table~\ref{tab:self}. 
A small self-BLEU score together with a large BLEU score can justify the effectiveness of a model, {\emph{i.e.}}, being able to generate realistic-looking as well as diverse samples. Among all the VAE variants, \emph{ml}-VAE-D shows the smallest BLEU score and largest unique $n$-grams percentage, demonstrating the effectiveness of hieararhically structured generative networks as well as latent variables. Even though AAE and ARAE yield better diversity according to both metrics, their corpus-level BLEU scores are much worse relative to \emph{ml}-VAE-D. We leverage human evaluation for further comparison.
\vspace{0mm}
%\begin{table}

%
\begin{table}
	\vspace{1mm}
	\def\arraystretch{1.2}
	\begin{center}
		\begin{small}
			\begin{tabular}{p{2.88in}}
				\toprule[1.2pt]
				we study the effect of disorder on the dynamics of a two-dimensional electron gas in a two-dimensional optical lattice , 
				we show that the superfluid phase is a phase transition , 
				we also show that ,
				in the presence of a magnetic field , 
				the vortex density is strongly enhanced . \\
				\hline
				in this work we study the dynamics of a colloidal suspension of frictionless ,
				the capillary forces are driven by the UNK UNK ,
				when the substrate is a thin film , 
				the system is driven by a periodic potential, 
				we also study the dynamics of the interface between the two different types of particles . \\
				%\hline
				%the problem of finding the largest partition function of a quantum system is equivalent to find the optimal value of the number of states ,
				%which is the most important example of the first order approximation of the ising model ,
				%the algorithm is based on the computation of the partition function ,
				%the results are compared with the results of the monte carlo simulations . \\
				\bottomrule[1pt]
			\end{tabular}
		\end{small}
	\end{center}
	\vspace{0mm}
	\caption{Generated arXiv abstracts from \emph{ml}-VAE-D model.}
	\label{tab:arxiv_hier_vae}
% 	\vspace{-3mm}
\end{table}

\begin{table}
	\centering
	%\vskip 0.0in
	\def\arraystretch{1.2}
	\setlength{\tabcolsep}{2.5pt}
	%\footnotesize 
	\small 
	\vspace{1mm}
	\begin{tabular}{cccccccc}
	   % \multicolumn{8}{c}{}\\
		\toprule[1pt]
		\multirow{2}{*}{\textbf{Model}} & \multicolumn{7}{c}{\textbf{Yelp}}\\
		\cmidrule(l){2-8} 
% 		&  	\textbf{\textcolor{azure(colorwheel)}{B-2}} & \textbf{\textcolor{azure(colorwheel)}{B-3}} & \textbf{\textcolor{azure(colorwheel)}{B-4}}  & \textbf{\textcolor{carrotorange}{2gr}} & \textbf{\textcolor{carrotorange}{3gr}} & \textbf{\textcolor{carrotorange}{4gr}}  & \textbf{\textcolor{ao(english)}{Etp-2}} \\
		&  	\textbf{B-2} & \textbf{B-3} & \textbf{B-4}  & \textbf{2gr} & \textbf{3gr} & \textbf{4gr}  & \textbf{Etp-2} \\
		\toprule[1pt]
		ARAE      &0.725 & 0.544 & 0.402 &36.2 & 59.7 & 75.8  & 7.551 \\
		AAE      & 0.831 & 0.672 & 0.483 &33.2 & 57.5 & 71.4  & 6.767\\
		\emph{flat}-VAE      & 0.872 & 0.755 & 0.617 & 23.7 & 48.2 & 69.0 & 6.793   \\
		\hline 
		\emph{ml}-VAE-S    & 0.865 & 0.734 & 0.591  & 28.7 & 50.4& 70.7 &6.843 \\
		\emph{ml}-VAE-D  & 0.851 & 0.723 & 0.579  & 30.5 & 53.2 & 72.6 & 6.926  \\
		\bottomrule[1.2pt]
		%\hline
	\end{tabular}
	\vspace{-2mm}
% 	\captionsetup{width=.90\textwidth}
	\caption{Evaluation of diversity of 1000 generated sentences on
\emph{self-BLEU scores} (\textbf{B-n}), \emph{unique $n$-gram percentages}  (\textbf{ngr}), \emph{$2$-gram entropy score}.}
% 	\emph{\textbf{\textcolor{carrotorange}{unique $n$-gram percentages}}} and \emph{\textbf{\textcolor{ao(english)}{$2$-gram entropy score}}}.}

% \emph{\textbf{\textcolor{azure(colorwheel)}{self-BLEU scores}}},
% 	\emph{\textbf{\textcolor{carrotorange}{unique $n$-gram percentages}}} and \emph{\textbf{\textcolor{ao(english)}{$2$-gram entropy score}}}.}
	%(smaller number indicates more diverse samples).
	\label{tab:self}
% 	\vspace{-2mm}
\end{table}
%\end{wraptable}

\noindent\textbf{Human Evaluation}
We conducted human evaluations using Amazon Mechanical Turk to assess the coherence and non-redundancy properties of our proposed models.  
Given a pair of generated reviews, the judges are asked to select their preferences (“no difference between the two reviews” is also an option) according to the following four evaluation criteria: 
\textit{fluency \& grammar}, \textit{consistency}, \textit{non-redundancy}, 
and \textit{overall}. We compare generated text from our \emph{ml}-VAE-D againts flat-VAE, AAE and real samples from the test set.  
Details are provided in the supplementary material.

As shown in Table~\ref{tab:human}, \textit{ml}-VAE generates superior human-looking samples compared to
\textit{flat}-VAE on the Yelp Reviews dataset. Even though both models underperform when compared against the ground-truth real reviews, \textit{ml}-VAE was rated higher in comparison to \textit{flat}-VAE (raters find \textit{ml}-VAE closer to human-generated than the \textit{flat}-VAE) in all the criteria evaluation criteria. When compared against AAE baseline models using the same data preprocessing steps and hyperparameters, \textit{ml}-VAE again produces more grammatically-correct and semantically-coherent samples. The human evaluations correlate with the automatic metrics, which indicate that our \textit{ml}-VAE is actually generating more coherent stories than the baseline models. We leave further evaluations using embedding based metrics as a possible extension to our work.
% \vspace{-2mm}
\subsection{Conditional Paragraph Generation}
% \vspace{-2mm}
% We further evaluate the proposed VAE model on a conditional generation task. 
We consider the task of generating an abstract of a paper based on the corresponding title.
The same arXiv dataset is utilized, where when training the title and abstract are given as paired text sequences.
The title is used as input of the inference network. For the generative network, instead of reconstructing the same input (\emph{i.e.}, title), the paper abstract is employed as the target for decoding.
We compare the \emph{ml}-VAE-D model against \emph{ml}-LM.
We observe that the \emph{ml}-VAE-D model achieves a test perplexity of $55.7$ (with a KL term of $2.57$), smaller that the test perplexity of \emph{ml}-LM ($58.1$), indicating that the information from the title is used by the generative network to facilitate the decoding process. Generated abstract samples from \emph{ml}-VAE-D model are shown in Table~\ref{tab:conditional_arxiv}. 

\begin{table}
	\begin{center}
		\vspace{2mm}
		\def\arraystretch{1.0}
		\begin{small}
			\begin{tabular} {p{2.8in}}%| p{2.5in}}
				\toprule[1.2pt]
				\textbf{Title}: \emph{Magnetic quantum phase transitions of the antiferromagnetic - Heisenberg model} \\
				\hline
				We study the phase diagram of the model in the presence of a magnetic field, The model is based on the action of the Polyakov loop, We show that the model is consistent with the results of the first order perturbation theory. \\ 
				\midrule[1.1pt]
				\textbf{Title}:  \emph{Kalman Filtering With UNK Over Wireless UNK Channels}  \\
				\hline
				The Kalman filter is a powerful tool for the analysis of quantum information, which is a key component of quantum information processing, However, the efficiency of the proposed scheme is not well understood .  \\
				\bottomrule[1.2pt]
			\end{tabular}
		\end{small}
		\vspace{-2mm}
		\caption{Conditionally generated arXiv paper abstracts from \emph{ml}-VAE-D model based on a given title. }
		\label{tab:conditional_arxiv}
		\vspace{-2mm}
	\end{center}
\end{table}

\begin{comment}
\begin{table}
	\centering
	%\vskip 0.0in
	\def\arraystretch{1.2}
	\vspace{2mm}
	\footnotesize 
	%\begin{scriptsize}
	\begin{tabular}{ccccc}
		%\hline
		\toprule[1.2pt]
		\textbf{Encoder Networks} & \textbf{NLL} & \textbf{KL} & \textbf{PPL}  \\
		\hline
		flat CNN encoder    & 164.6   &   2.3   &    50.2     \\
		multi-level LSTM encoder      & 161.3&	 5.7     &  46.9   \\
		hierarchical CNN encoder      &  160.2   &   6.8  &     45.8 \\
		\bottomrule[1.2pt]
		%\hline
	\end{tabular}
	\vspace{-2mm}
	\caption{Ablation study with different encoders.}
	\label{tab:encoder}
	%\end{scriptsize}
	\vspace{3mm}
\end{table}
\end{comment}
\begin{table}
	%\begin{wraptable}{L}{2.0in}
	\centering
	%\vskip 0.0in
	\def\arraystretch{1.2}
	\setlength{\tabcolsep}{2.8pt}
	\vspace{2mm}
	\small  
	%\begin{scriptsize}
	\begin{tabular}{ccccc}
		%\hline
		\toprule[1.2pt]
% 		\textbf{Model} &  	\textbf{\textcolor{azure(colorwheel)}{G}} & \textbf{\textcolor{carrotorange}{C}} & \textbf{\textcolor{ao(english)}{Non-R}}  & \textbf{\textcolor{mypurple}{Overall}} \\
		\textbf{Model} &   \textbf{Grammar.} & \textbf{Cons.} & \textbf{Non-Red.}  & \textbf{Overall} \\
		\hline
		%LeakGAN      & 0.8574 & 0.6963 & 0.5532  \\
		%\emph{flat}-VAE (w/o $z_2$)      & - & - & - & -    \\
		\emph{ml}-VAE    & 52.0 & 55.0 & 53.7 & 60.0    \\
		\emph{flat}-VAE      & 30.0 & 33.0 & 27.7 & 32.3  \\
		\hline 
		\emph{ml}-VAE  & 75.3 & 86.0 & 76.7 & 86.0    \\
		AAE  & 13.3 & 10.3 & 15.0 &  12.0  \\
		\hline 
		\emph{flat}-VAE  & 19.7 & 18.7 & 14.3 & 19.0    \\
		Real data  & 61.7 & 74.7 & 74.3 & 77.7    \\
		\hline 
		\emph{ml}-VAE  & 28.0 & 26.3 & 25.0 & 30.3    \\
		Real data  & 48.6 & 58.7 &  49.0 & 61.3  \\
		\bottomrule[1.2pt]
		%\hline
	\end{tabular}
	\vspace{-2mm}
	\caption{Human evaluations on Yelp Reviews dataset. Each block is a head-to-head comparison of two models on grammatically, consistency, and non-redundancy. 
		%From columns 2-5: Grammaticality(\textbf{G}), Consistency (\textbf{C}), Non-Redundancy(\textbf{N}), Overall (\textbf{O}). 
	}
	\label{tab:human}
	%\end{scriptsize}
 	\vspace{2mm}
	%\end{table}
\end{table}
\vspace{-1mm}
% ###############################
\subsection{Analysis}
% ###############################
\begin{comment}
\subsubsection{Ablation Study}
\vspace{-1mm}
%\subsubsection{The architecture of encoder networks}
To investigate the impact of encoder networks on the VAE's performance, we conduct an ablation study, where the hierarchical CNN encoder in the \emph{ml}-VAE model is replaced with a flat CNN encoder or a multi-level LSTM encoder. The corresponding results are shown in Table~\ref{tab:encoder}. It is observed that the model with a flat CNN encoder yields worst (largest) perplexity, suggesting that it is beneficial to make the encoder hierarchical. Additionally, the hierarchical CNN encoder exhibits slightly better results than multi-level LSTM encoder, according to our experiments.
\end{comment}
\vspace{-1mm}
\paragraph{The Continuity of Latent Space}
Following \citep{Bowman2016GeneratingSF}, we measure the continuity of the learned latent space. We randomly sample two points from the prior latent space (denoted as $A$ and $B$) and generate sentences based on the equidistant intermediate points along the linear trajectory between $A$ and $B$. 
As shown in Table~\ref{tab:interpolate}, these intermediate samples are all realistic-looking reviews that are syntactically and semantically reasonable, demonstrating the smoothness of the learned VAE latent space. 
Interestingly, we even observe that the generated sentences gradually transit from positive to negative sentiment along the linear trajectory.
To validate that the sentences are not generated by simply retrieving the training data, we find the closest instance, among the entire training set, for each generated review. Details of the results can be found in the supplementary material (Table~\ref{tab:interpolate_full}).
\begin{table}
	\begin{center}
		\vspace{-1mm}
		\def\arraystretch{1.2}
		\setlength{\tabcolsep}{3pt}
		\begin{small}
			\begin{tabular} {c p{2.8in}}
				\toprule[1.2pt]
				%\midrule[1.1pt]
				\textbf{A} & the service was great, the receptionist was very friendly and the place was clean, we waited for a while, and then our room was ready . \\ 
				\hline
				$\bullet$ &  same with all the other reviews, this place is a good place to eat, i came here with a group of friends for a birthday dinner, we were hungry and decided to try it, we were seated promptly.  \\
				\hline
				$\bullet$ &  this place is a little bit of a drive from the strip, my husband and i were looking for a place to eat, all the food was good, the only thing i didn t like was the sweet potato fries.  \\
				\hline
				$\bullet$ &  this is not a good place to go, the guy at the front desk was rude and unprofessional, it s a very small room, and the place was not clean. \\
				\hline 
				$\bullet$ &  service was poor, the food is terrible, when i asked for a refill on my drink, no one even acknowledged me, they are so rude and unprofessional. \\
				\hline
				\textbf{B} &  how is this place still in business, the staff is rude, no one knows what they are doing, they lost my business .  \\
				\bottomrule[1.2pt]
			\end{tabular}
		\end{small}
		\vspace{-2mm}
		\caption{
			Intermediate sentences are produced from linear transition between two points in the latent space and sending them to the generator network.
		} 
		\label{tab:interpolate}
		\vspace{-2mm}
	\end{center}
	%\end{table}
\end{table}
\vspace{-2mm}
\paragraph{Attribute Vector Arithmetic}
%To investigate the structure of the latent space, 
We conduct an experiment to alter the sentiments of reviews with an \emph{attribute vector}.
We encode the reviews of the Yelp Review training dataset with positive sentiment and sample a latent code for each review and measure the mean latent vector.  
The mean latent vector of the negative reviews are computed in the same way. We subtract the negative mean vector from the positive mean vector to obtain the ``sentiment attribute vector". Next, for evaluation, we randomly sample $1000$ reviews with negative sentiment and add the ``sentiment attribute vector" to their latent codes. The manipulated latent vectors are then fed to the hierarchical decoder to produce the transferred sentences, hypothesizing that they will convey positive sentiment. 

As shown in Table~\ref{tab:transfer}, the original sentences have been successfully manipulated to positive sentiment with the simple attribute vector operation. 
However, the specific contents of the reviews are not fully retained. One interesting future direction is to decouple the style and content of long-form texts to allow \emph{content-preserving} attribute manipulation.
We employed a CNN sentiment classifier to evaluate the sentiment of manipulated sentences. The classifier is trained on the entire training set and achieves a test accuracy of $94.2\%$. With this pre-trained classifier, $83.4\%$ of the transferred reviews are predicted as positive-sentiment, indicating that ``attribute vector arithmetic'' consistently produces the intended manipulation of sentiment. 
\begin{table}
	\begin{center}
		\vspace{1mm}
		\def\arraystretch{1.2}
		\begin{small}
			\begin{tabular} {p{2.7in}} %p{2.6in}}
				\toprule[1.2pt]
				\textbf{Original}: you have no idea how badly i want to like this place, they are incredibly vegetarian vegan friendly , i just haven t been impressed by anything i ve ordered there , even the chips and salsa aren t terribly good , i do like the bar they have great sangria but that s about it .    \\
				\hline 
				\textbf{Transferred}:  this is definitely one of my favorite places to eat in vegas , they are very friendly and the food is always fresh, i highly recommend the pork belly , everything else is also very delicious, i do like the fact that they have a great selection of salads .
				\\
				\bottomrule[1.2pt]
			\end{tabular}
		\end{small}
		\vspace{-2mm}
		\caption{An example sentiment transfer result with attribute vector arithmetic. More examples can be found in the supplementary material (Table~\ref{tab:transfer_add}).}
		\label{tab:transfer}
		\vspace{-2mm}
	\end{center}
\end{table}

\section{Conclusion}
We introduce a hierarchically-structured variational autoencoder for long text generation. It consists of a multi-level LSTM generative network to model the semantic coherence at both the word- and sentence-levels. A hierarchy of stochastic layers is employed, where the priors of the latent variables are learned from the data. Consequently, more informative latent codes are manifested, and the generated samples also exhibit superior quality relative to those from several baseline methods.
%(according to both automatic metrics and human evaluation). 
%Human evaluations demonstrate that the samples from \emph{ml}-VAE-D are less repetitive and semantically-consistent.

\bibliography{acl2019}
\bibliographystyle{acl_natbib}

\end{document}